\documentclass{article}
\usepackage{spconf,amsmath,graphicx,amssymb,epstopdf,algorithm,color,}
\usepackage{tabularx}
\usepackage{algorithm}
\usepackage{caption}

\usepackage[noend]{algpseudocode}
\algnewcommand\algorithmicinput{\textbf{Input:}}
\algnewcommand\INPUT{\item[\algorithmicinput]}
\algnewcommand\algorithmicoutput{\textbf{Output:}}
\algnewcommand\OUTPUT{\item[\algorithmicoutput]}

\def\R{{\mathbb{R}}}

\title{A Searchlight Factor Model Approach for Locating Shared Information in Multi-Subject fMRI Analysis}

\name{\begin{tabular}{c}Hejia Zhang${}^{1}$, Po-Hsuan Chen${}^1$, Janice Chen${}^{2}$, Xia Zhu${}^{3}$,  Javier S. Turek${}^{3}$,\\
Theodore L. Willke${}^{3}$, Uri Hasson${}^{2}$, Peter J. Ramadge${}^{1}$\end{tabular}}

\address{${}^1$Department of Electrical Engineering, Princeton University, \\
${}^2$Princeton Neuroscience Institute and Department of Psychology, Princeton University, ${}^3$Intel Labs\\
}

\begin{document}
%\ninept
%
\maketitle
\begin{abstract}
%Because of the scarcity of fMRI data from a single subject, 
There is a growing interest in joint multi-subject fMRI analysis. The challenge of such analysis comes from inherent anatomical and functional variability across subjects. One approach to resolving this is a {\em shared response factor model}. 
This assumes a shared and time synchronized stimulus across subjects.
%However, it suffers from the lack of spatial locality. That is, it 
Such a model can often identify shared information, but it may not be able to pinpoint with high resolution the spatial location of this information. In this work, we examine a searchlight based shared response model to identify shared information in small contiguous regions (searchlights) across the whole brain. Validation using classification tasks 
demonstrates that we can pinpoint informative local regions.

\end{abstract}
\section{Introduction}
\label{sec:intro}
%\subsection{Background}
%Since a limited amount of fMRI data can be gathered from a subject in a single session,
The sensitivity of the statistical data analysis of fMRI data is limited by amount of data available. 
%that can be gathered from a subject in a single session
Hence there is an increasing interest in gathering data from multiple sessions and/or multiple subjects. It is hoped that this will increase statistical sensitivity in testing hypotheses about human cognition \cite{friston1999multisubject,mumford2006modeling}. 
Here we focus on the problem of analyzing fMRI data across multiple subjects
all experiencing the same time synchronized stimulus. 
The raw fMRI data is thus temporally aligned, but is
neither anatomically nor functionally aligned across subjects.
Traditional methods of addressing this via anatomical registration, \cite{talairach1988co,fischl1999high,mazziotta2001probabilistic}, 
ignore the inherent functional variability across subjects. 
More recent approaches attempt to use anatomical and functional data to align, or register, functional structure \cite{conroy2009fmri,sabuncu2010function,conroy2013inter}. 
A distinct approach learns a factor model that jointly models the functional responses.  
Such factor methods have shown promising results and are attracting increased attention, e.g.,  \cite{WICS:WICS101,hyvarinen2004independent,lee2008independent,calhoun2001method,haxby2011common,chen2015srm}.
%\subsection{Factor Models}

To make things more precise let
$X_i \in \mathbb{R}^{v_x\times v_y\times v_z\times t}$ denote the data from subject $i$,  
$i = 1:m$, where $(v_x, v_y, v_z)$ are the number of voxels in the 3D volume along the $(x,y,z)$ axes, and $t$ is number of time samples (TRs) in the experiment. 
Typically one re-arranges $X_i$ into a $v$ by $t$ matrix, where $v=v_x\times v_y\times v_z$. 
If one is only interested in voxels within a given region of interest (ROI), then $X_i$ is simply the sub-matrix formed by restriction to those voxels.
In this setting, the basic goal is to jointly factorize the matrices $X_i$ into a product 
$W_i S$ so as to minimize some constrained or regularized objective function. 
Here 
$k$ is the number of factors,
$W_i \in \mathbb{R}^{v\times k}$,  
and $S \in \mathbb{R}^{k\times t}$. 
The ``mixing matrices'' $W_i$ are subject specific (acknowledging the distinct functional topographies of the subjects), while the ``source matrix'' $S$ provides a set of $k$ shared elicited time responses. 
For this reason we call such models {\em shared response models}. 

Form $X = [X_1^T\dots X_m^T]^T \in \R^{mv\times t}$ by stacking the $X_i$ along the spatial dimension, and similarly form 
$W = [W_1^T\dots W_m^T]^T \in \R^{mv\times k}$.
Then $X =WS + E$ where $S\in \R^{k\times t}$ is the shared response and $E$  is the model error. $W$ and $S$ are then determined by minimizing a regularized cost function 
of the terms in the decomposition subject to constraints on $W$ and $S$.

As a specific example, PCA  selects $W$ and $S$ to minimize $\|X - W S\|_F^2$.
The resulting $W$ has orthonormal columns and is partitioned into $m$ sub-matrices $W_i$, $i=1:m$. 
 %The $W_i$ need not have ON columns. 
Since $S = W^T X = \sum_i W_i^T X_i$, the sources are formed as linear
combinations of potentially all voxel time courses.
%This leads to unnatural joint orthogonality across $W_i$.
Similarly, using a compact SVD: $X= U\Sigma V^T$, $U = XV\Sigma^{-1}$, and
$W$ is the first $k$ columns of $U$. So the columns of $W$ are 
linear combinations of potentially all columns of $X$. 

In general, in a shared response factor model $W_i$ can be formed as a linear combination of all columns in $X_i$ and $S$ can be formed using a linear combination of all voxel time courses. In this sense the factor model has no assurance of preserving spatial locality.
Such methods are usually applied to large pre-defined ROIs, such as PMC \cite{chen2015srm} or ventral temporal cortex (VT) \cite{haxby2011common}, to investigate the relationship between the ROI and a specific stimulus or cognitive state. 
%However, the ROI-based factor models also suffer from the lack of spatial locality, so they are unable to find the more fine-grained functional structure in the large ROI. Therefore, a way to preserve spatial locality in factor models is needed to make more meaningful neuroscience inferences.

%\begin{figure}[!tb]
%\centering
%\includegraphics[width=0.8\linewidth]{dispersion_preliminary.pdf}%
%\caption{\small Illustration of the lack of spatial locality of whole brain SRM analysis. 
%The support of $M\prime $ is not close to that of $M$.
%}\label{fig:sploc1}\vspace{-1em}\end{figure}

%\subsection{Searchlight Analysis}
In this paper we specifically set out to preserve spatial locality in forming the shared response factor model.
This is important in exploratory analyses where one seeks to detect and study 
small local regions in the brain where shared information is present.
There are two possible approaches: a searchlight analysis 
\cite{kriegeskorte2006information,etzel2013searchlight, guntupalli2016model,chen2016convolutional}, 
or adding spatial regularization to the factorization cost function. 
%when learning the shared response model. 
Both approaches are worthy of study. Here we focus on the searchlight approach.

\vspace{-0.55mm}
A searchlight uses a fixed number of neighboring voxels to conduct analysis for each voxel location, and the same analysis is conducted on all locations. This idea can be used to extend any ROI-based factor model to overlapping searchlights.  
We combine factor models of the SRM form and searchlight analysis to enable localized analysis in the whole brain multi-subject fMRI functional alignment. In detail, a fixed sized searchlight centering at voxel $i$ is used to scan over the whole brain. For each searchlight location, a factor model is used to functionally align across subjects, and an analysis is performed based on the results of the alignment. Statistics from the analysis 
(e.g., classification accuracy) is assigned to the center voxel $i$ of that specific searchlight. 
We report the accuracy of each searchlight on a given classification task.
This helps identify locations in the brain in which information is shared across subjects for a specific cognitive task. 

\vspace{-0.55mm}
For multi-subject neuroscience datasets and experiments, we provide an effective method for locating where the shared information is over the whole brain while keeping the quality of the found shared information. So our method can serve as a first step in multi-subject fMRI analysis to help identify regions that worth further investigation in a neuroscience experiment. In some factor models, the number of latent factors $k$ can be pre-specified. In this case, we record the k value that gives the best analysis result 
%(e.g. highest accuracy) 
for each searchlight.
% (Fig.~\ref{fig:acc_k}). 
We report both the accuracy and the best $k$ value on a brain map as a proxy for the presence and richness of a shared cognitive state across subjects.
We also developed two new variants of the ICA and group ICA factor models. These variants show good performance in the functional alignment task. 

%\begin{figure}[!tb]
%\centering
%\includegraphics[width=0.8\linewidth]{accu_vs_k.pdf}%
%\caption{\small Classification accuracy vs. number of factors k in SRM for a single searchlight.
%}\label{fig:acc_k}\vspace{-1em}
%\end{figure}

\begin{algorithm}[t]
\caption{Shared Response ICA (SR-ICA)}\label{alg:sr-ica}
  \begin{algorithmic}[1]
    \INPUT Data matrices $X_i$, number of factors $k$, convergence threshold $\tau$, max iteration $N$, number of subjects $m$
		\State $W_i^0 \gets $ initialization with random orthonormal columns 
		\For{$n$ in $1$ to $N$}
			\State $S \gets \frac{1}{m} \sum _{i=1}^m {W_i^{n-1}}^+ X_i$ \Comment{$(\cdot)^+$ is pseudo-inverse}
			\For{$i$ in $1$ to $m$}
				\State $W_i^{n} \gets (E\lbrace X_i g(S)\rbrace -E\lbrace X_i g' (S)\rbrace {W_i^{n-1}}^+)^+$
				\State $W_i^{n} \gets W_i^{n} ({W_i^{n}}^T W_i^{n})^{-1/2} $
			\EndFor
			\State converged $\gets True$
			\For{$i$ in $1$ to $m$}
				\If{max$\vert {W_i^{n}}^T W_i^{n-1} -I\vert \geq \tau $}
					\State converged $\gets False$
				\EndIf
			\EndFor
			\If{converged}
				 \textbf{break}
			\EndIf
      	\EndFor
     	\State \Return{$W_i,S$}      
	%\OUTPUT Subject-specific maps $W_i$ and shared response $S$            
  \end{algorithmic}
\vspace{-1mm}

\end{algorithm}
\begin{algorithm}[t!]
 \caption{Shared Response Group ICA (SR-GICA)}\label{alg:GICA}
  \begin{algorithmic}[1]
    \INPUT Data matrices $X_i$, number of factors $k_1,k_2$, number of subjects $m$
		\For{$i$ in $1$ to $m$}
			\State $X_i = F_iP_i$ \Comment{First PCA with $k_1$ components} 
		\EndFor
		\State $P \gets [P_1^T,\dots, P_m^T]^T $
		\State $P = GY$ \Comment{Second PCA with $k_2$ components}
		\State $Y = AS$ \Comment{ICA with $k_2$ components}
		\State Partition $[G_1^T, \dots, G_m^T]^T \gets G$
		\State Then, $G_iAS=P_i \rightarrow F_iG_iAS=F_iP_i=X_i$ 
		\State $W_i \gets F_iG_iA$
     	\State \Return{$W_i,S$}      
	%\OUTPUT Subject-specific maps $W_i$ and shared response $S$            
  \end{algorithmic}
\end{algorithm}

\vspace{-6mm}
\section{Some Specific Factor Models} \label{sec:factormodel}

\vspace{-2mm}
Examples of  (probabilistic and deterministic) factor methods that have been used in multi-subject fMRI analysis include 
PCA \cite{WICS:WICS101}, 
ICA \cite{hyvarinen2004independent,lee2008independent}, 
Group ICA \cite{calhoun2001method}, 
hyperalignment (HA) \cite{haxby2011common}, 
and shared response model (SRM) \cite{chen2015srm,anderson2016enabling}. 

The desirable factor models to be combined with searchlight analysis in multi-subject functional alignment should have the following properties: 1) has an adjustable number of factors $k$; 2) shows good performance in large area multi-subject functional alignment. Below, we discuss and compare several candidates using a consistent framework. 
%$X_i \in \R^{v\times t}$ is the data matrix of subject i. By stacking $X_i$ along the spatial dimension we get $X = [X_1^T\dots X_m^T]^T \in \R^{mv\times t}$ . $W_i \in \R^{v\times k}$ is the functional topographies of subject i. Stacking $W_i$ along the spatial dimension gives us $W = [W_1^T\dots W_m^T]^T \in \R^{mv\times k}$. $S\in \R^{k\times t}$ is the shared response. 

%\subsection{PCA, HA, SRM}

\vspace{1mm}
\noindent {\em PCA:}
%PCA maximizes the variance in the projected data \cite{WICS:WICS101}. 
Standard PCA is a deterministic model. We have already
outlined how to obtain the $W_i$ and $S$ in this case.
%There are two ways to apply PCA to multi-subject data, on temporally stacked joint data matrix or spatially stacked joint data matrix. Here we explore spatially stacked approach due to assumption of temporally synchronized response, the other approach has been discussed in \cite{chen2014joint}. 
%To do so PCA minimizes $\|X - W S\|_F^2$.The resulting $W$ has orthonormal (ON) columns.
% $W$ is then partitioned into $m$ sub-matrices $W_i$, $i=1:m$. 
% %The $W_i$ need not have ON columns. 
%Since $S = W^T X = \sum_i W_i^T X_i$, the sources are formed as linear
%combinations of potentially all voxel time courses.
%%This leads to unnatural joint orthogonality across $W_i$.
%Similarly, using a compact SVD we can write $X= U\Sigma V^T$, $U = XV\Sigma^{-1}$, and
%$W$ is the first $k$ columns of $U$. So the columns of $W$ are 
%linear combinations of potentially all of the columns of $X$. 

\vspace{1mm}
\noindent {\em SRM \cite{chen2015srm}:}
The shared response model in \cite{chen2015srm} minimizes 

\vspace{-3mm}
\begin{equation}\label{eq:Fcost}
\textstyle \sum_{i=1}^m \frac{1}{m}\|X_{i} - W_{i} S\|_F^2,
\vspace{-1mm}
\end{equation} 
 subject to $W_i^TW_i=I_k$.
This is done in a graphical model framework that provides Bayesian regularization.
This SRM can be considered as a variant of pPCA in a multi-source setting \cite{ahn2003constrained,chen2015srm}.
%$W_i$ is the subject-specific functional topography mappings and only has $k$ columns, where
%In SRM, $k$ is smaller than $v$ in most of the cases. $S$ is considered as the shared response across all subjects. 
The shared response of some held out data $X_i^\prime$ for subject $i$ can be computed as $S_i^\prime  = W_{i}^T X_{i}^\prime$. 

\vspace{1mm}
\noindent {\em Hyperalignment (HA) \cite{haxby2011common}:} 
HA is a deterministic model that learns orthogonal $W_i\in \R^{v\times v}$ to minimize (\ref{eq:Fcost}). It does not have a selectable factor dimension $k<v$.
For $k=v$, the non-probabilistic version of the SRM in \cite{chen2015srm} yields HA.  
So HA can be considered a special case of the SRM in \cite{chen2015srm}.
A searchlight application of HA has been explored in \cite{guntupalli2016model}. 
For these reasons we do not explore HA further here.

\vspace{1mm}
\noindent {\em Independent Component Analysis (ICA):}
ICA learns statistically independent signals as measured 
%minimize the gaussianity, or mutual information of the estimated signals as 
by kurtosis or negentropy \cite{hyvarinen2004independent,lee2008independent}.
We use the FastICA algorithm, an efficient probabilistic method \cite{hyvarinen2004independent} optimizing negentropy of the shared response $S$. This is formulated as \vspace{-0.5em}
\begin{equation}\label{eq:ICAcost}
\textstyle \max_W [E(G(S)) - E(G(\mathcal{N}))]^2,
\vspace{-0.5em}
\end{equation} 
where $G(\cdot)$ is a nonquadratic function, e.g. $\log\cosh$, and $\mathcal{N}$ is a standard normal random variable. 
This yields $X=WS+E$. $W$ is then partitioned into $m$ sub-matrices $W_i$, $i=1:m$. %This also leads to unnatural joint orthogonality across $W_i$.

\vspace{1mm}
\noindent {\em SR-ICA:}
We also study a new algorithm we call shared response ICA (SR-ICA). Motivated by the framework in \cite{chen2015srm}, we modify the FastICA algorithm.
In SR-ICA, the block structure of the subject data is preserved by spatial concatenation in both $X$ and $W$. The key difference is that instead of learning a joint matrix $W$, we iteratively learn $W_i$ to ensure block-wise structure in $W$. 
%rather than orthogonalizing all the rows of unmixing matrix $U$ in each iteration, we orthogonalize the rows of each submatrix $U_i$.
This is summarized in Algorithm \ref{alg:sr-ica}\footnote{We acknowledge the help of Jacob Simon in coding SR-ICA.}. Here we follow the convention of working with unmixing matrices $U_i$ instead of $W_i$. The function $g(\cdot)$ is the derivative of $G(\cdot)$ in \eqref{eq:ICAcost} \cite{hyvarinen2000independent}.

\vspace{1mm}
\noindent {\em Group ICA (GICA):}
Group ICA, an algorithm for making group inferences, uses two applications of PCA 
and an application of ICA \cite{calhoun2001method}. The original algorithm first performs subject specific PCA along the temporal dimension. Then the projected data matrices for all subjects are concatenated to form a joint data matrix. A second PCA is then performed on the joint data matrix. Lastly, an ICA is performed on the projected data matrix after the second PCA. 
%dimension reduction along the temporal dimension and concatenates reduced data matrices along this reduced dimension. 
%GICA has been shown to work well\cite{calhoun2001method}, however, it doesn't quite work in our problem due to its not utilizing the temporal synchronization property in our dataset.
%for data with a temporally synchronized stimulus, the reduction along temporal dimension throws away the synchronization. The original group ICA does not perform well in our experiments in its original setting. We utilize the property with GICA by 
We apply GICA along the spatial dimension to learn a low dimensional shared response space. See Algorithm \ref{alg:GICA}.

\begin{table}[t!b]
\vspace{0em}
\centering
\footnotesize
\begin{tabularx}{0.45\textwidth}{ ll }
{\bf Dataset}  & {\bf TRs (s/TR)}  \\
\hline
audiobook (narrated story) \cite{Yeshurun2014How}         	 & 449(2)  \\ 
sherlock-movie (audio-visual movie) \cite{chen2016shared}	     & 1976(2) \\      
sherlock-recall (movie free-recall) \cite{chen2016shared}	     & 437$\sim$1009(2)\\     
\end{tabularx}
\vspace{0em} 
\captionof{table}{ \small fMRI datasets used in the experiments. All datasets are whole brain (WB) in MNI \cite{mazziotta2001probabilistic}. We use 9 subjects in each dataset. } \label{tab:dataset} 
\vspace{-2em}
\end{table}

\begin{figure}[t!b]
\centering
\includegraphics[width=1\linewidth]{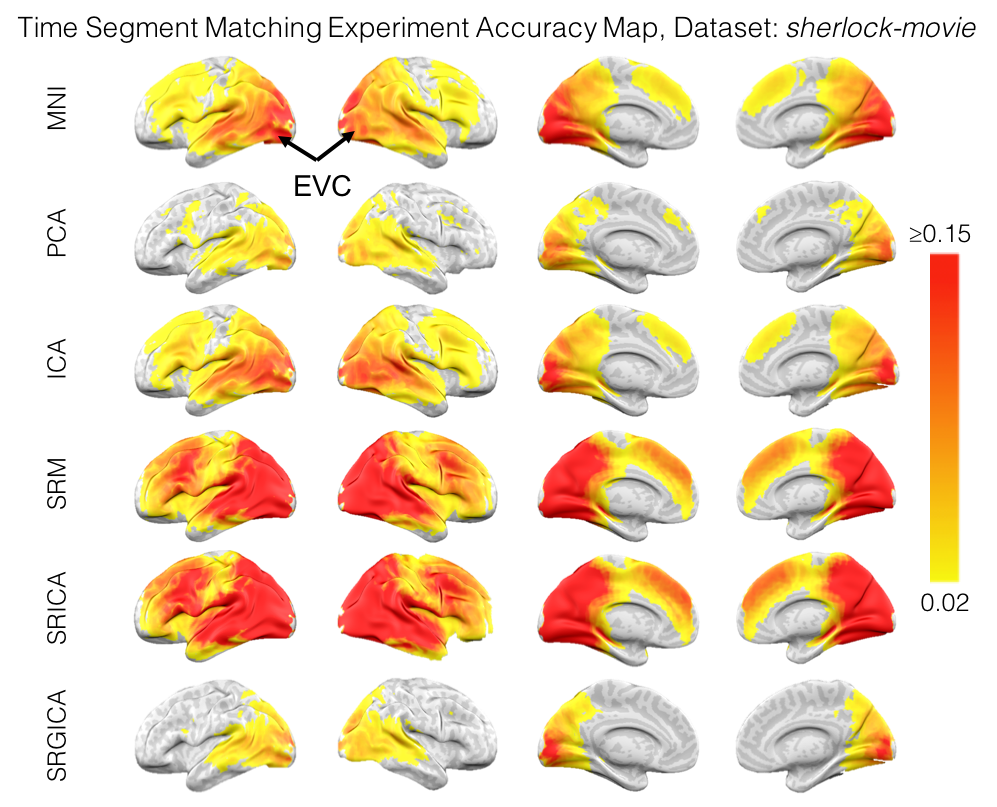}%
%\caption{\small Accuracy brain maps for time segment matching using $sherlock$-$movie$ dataset.
%}\label{fig:sherlock_acc}\vspace{-1em}
%\end{figure}

%\begin{figure}[!tb]
%\centering
\includegraphics[width=1\linewidth]{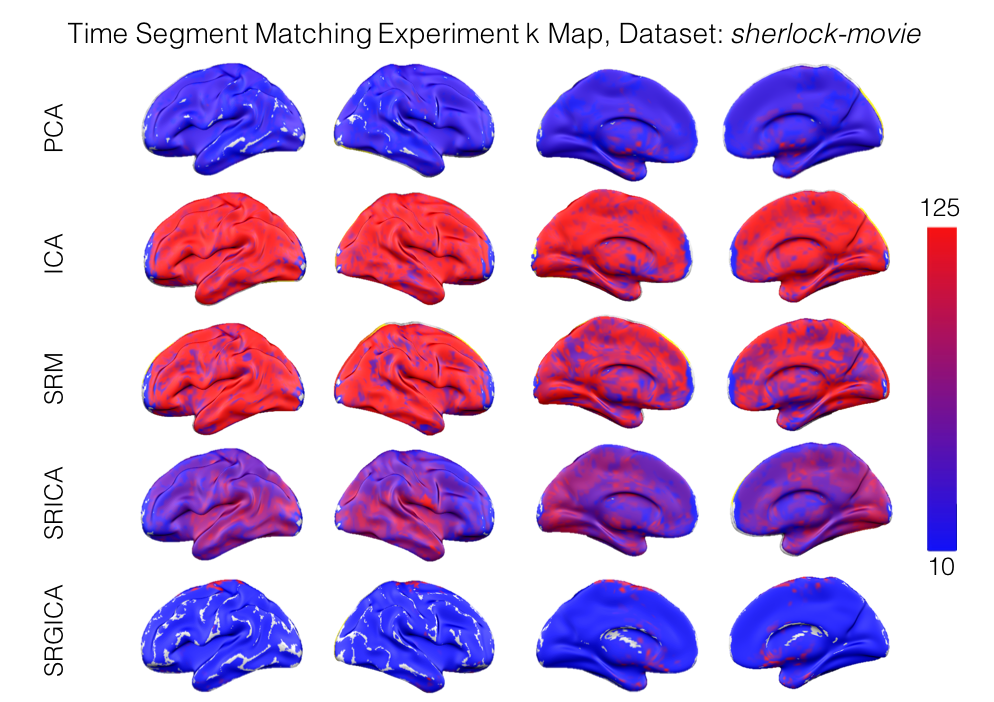}%
\caption{\small 
Top: Accuracy and $k$ brain maps for time segment matching using $sherlock$-$movie$ dataset. EVC: early visual cortex.
%Bottom: ``Best'' $k$ brain maps for time segment matching using $sherlock$-$movie$ dataset. ``Best'' $k$ gives highest accuracy for that searchlight.
}\label{fig:sherlock_acc_k}

\vspace{-6mm}
\end{figure}

\vspace{-2mm}
\section{Experiments and Results} \label{sec:experiments}
Three datasets (Table \ref{tab:dataset}),
collected using a 3T Siemens Skyra scanner with different preprocessing pipelines,
are used to test and compare the searchlight factor models. 
 The {\em sherlock-movie} dataset was collected while the subjects watched a $50$ minute BBC episode "Sherlock". The {\em sherlock-recall} dataset was collected while the same subjects verbally reiterated the "Sherlock" episode without any hints. The {\em audiobook} dataset was collected while a different set of subjects listened to a $15$ minute narrated story.

In all experiments, we use $5 \times 5 \times 5$ searchlights on data down-sampled by 2. For each searchlight, we try $k=[10,25,50,75,100,125]$ and report the highest accuracy and the corresponding $k$ value. To ensure a fair comparison, for SR-GICA, we set $k_1$ to be the number of voxels in the searchlight and $k_2=k$. We tried other $k_1$ values, but this resulted in less accuracy.  Note that there are relatively few voxels per searchlight to begin with. 
The accuracies are computed based on the projected shared response of held out data using learned subject-specific maps  $W_i\in \mathbb{R} ^{125\times k}$. 

\begin{figure}[!tb]
\centering
\includegraphics[width=1\linewidth]{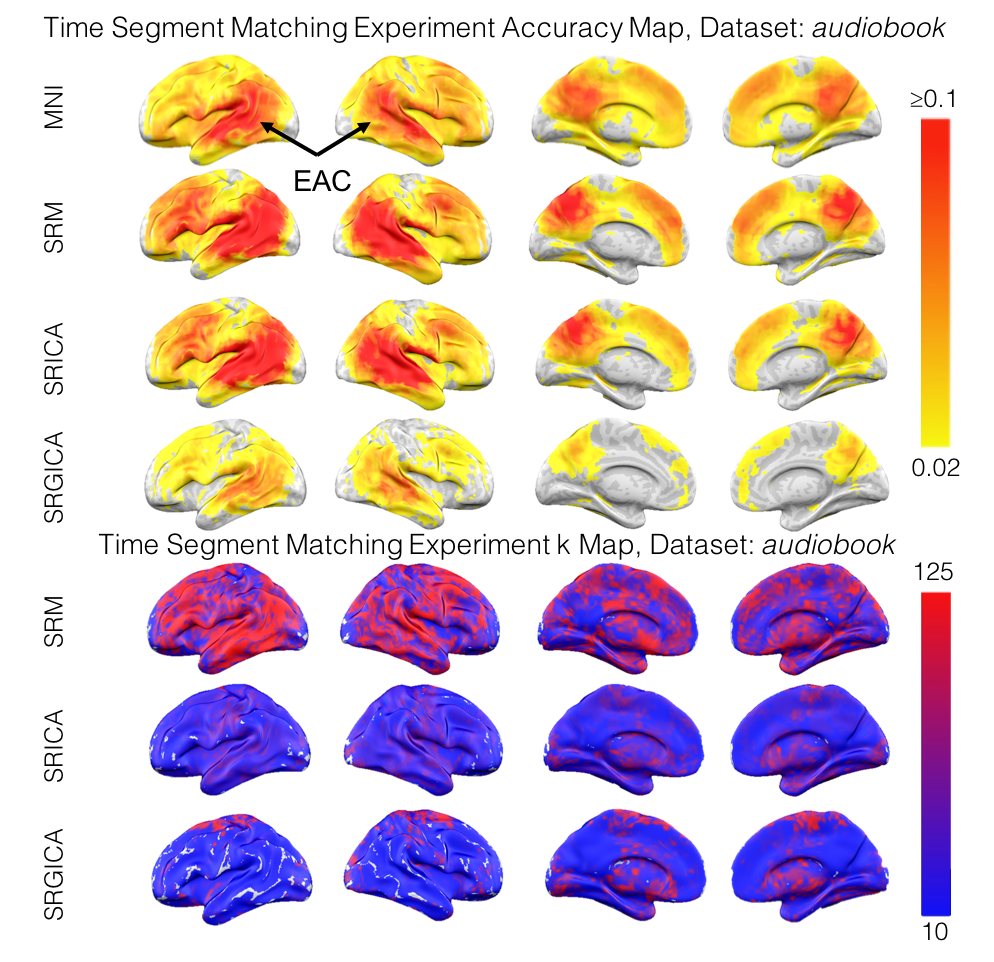}%
\caption{\small Accuracy and k brain maps for time segment matching experiment using audiobook dataset. EAC: early auditory cortex.
}\label{fig:greeneye_acc_k}

\vspace{-4mm}
\end{figure}

\begin{figure}[!tb]
\centering
\includegraphics[width=1\linewidth]{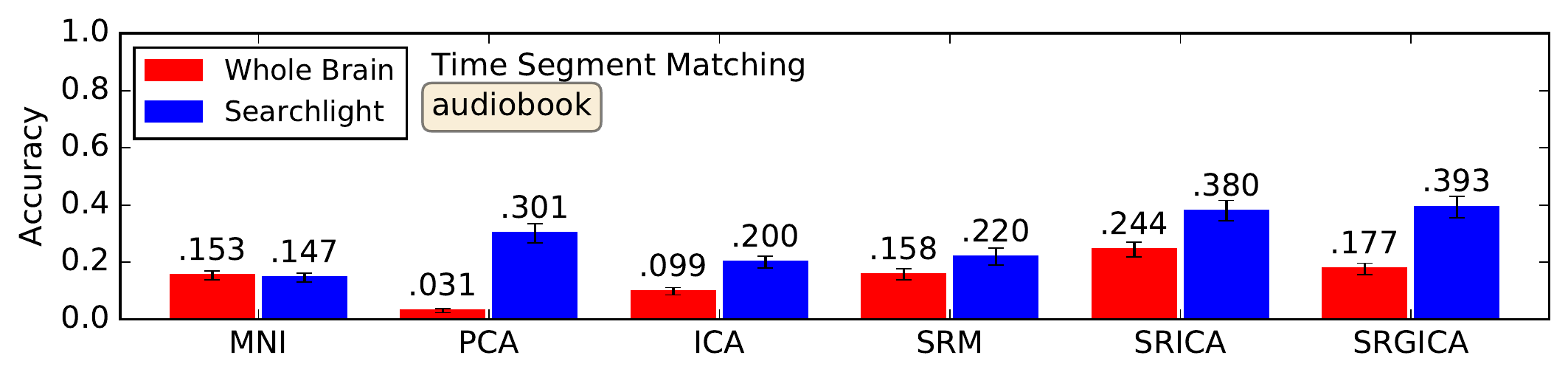}
%\caption{\small Time segment matching accuracy for audiobook dataset. WB accuracies in red and Searchlight accuracies in blue.
%}\label{fig:greeneye_bar}

\vspace{-1.5mm}
%\end{figure}
%\begin{figure}[!tb]
%\centering
\includegraphics[width=1\linewidth]{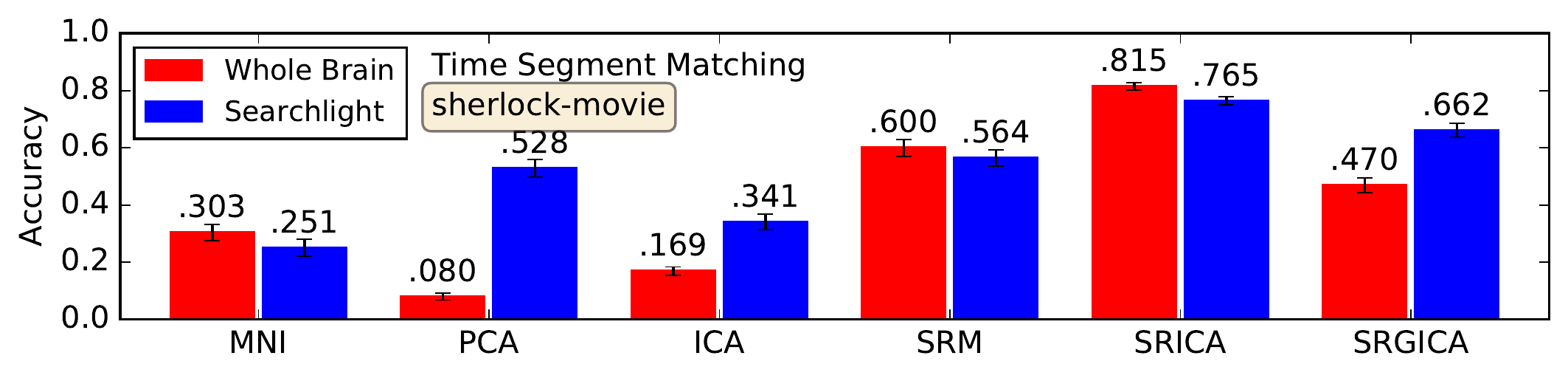}
%\caption{\small Time segment matching accuracy for $sherlock$-$movie$ dataset. WB accuracies in red and Searchlight accuracies in blue.
%}\label{fig:sherlock_bar}\vspace{-1em}
%\end{figure}
%\begin{figure}[!tb]
%\centering

\vspace{-1.5mm}
\includegraphics[width=1\linewidth]{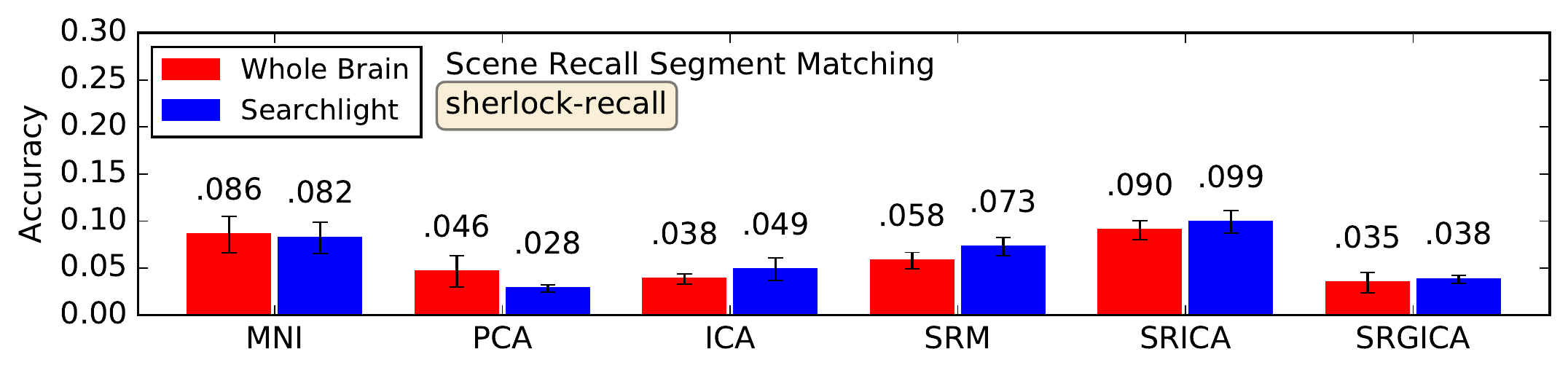}
\caption{\small 
Top and Middle:
Time segment matching accuracies (top: audiobook, middle: sherlock-movie). 
Bottom:
Scene recall matching accuracy (sherlock-movie). 
%WB in red and Searchlight in blue.
}\label{fig:acc_bar}\vspace{-6mm}
\end{figure}

\vspace{1mm}
\noindent{\em Time Segment Matching:}
This experiment is designed to test if the shared response we learned can be generalized to new data. That is, what is the quality of the shared information extracted. We use the {\em audiobook} and {\em sherlock-movie} datasets. The fMRI data are split into two halves along the temporal axis, one for training and the other for testing, and the roles reversed and the results averaged. In this experiment, we first use training data of all 9 subjects for learning the shared response. Then, a random 9 TR time segment from the testing subject (1 of the 9 subjects), called test segment, projected to the shared response space. The other 8 subjects' testing data is projected to the shared response space and averaged.  We then locate this time segment by maximizing Pearson correlation between the average response and response from the test segment. Segments overlapping with the test segment are excluded in matching. 
Assuming independent time segments, chance accuracy is $0.0044$ for {\em audiobook} and $0.001$ for {\em sherlock-movie}. The accuracy and k brain maps for different searchlight factor models are shown in 
%using {\em sherlock-movie} dataset 
in Fig.~\ref{fig:sherlock_acc_k}
%. The brain maps for some models using {\em audiobook}  dataset is in 
and Fig.~\ref{fig:greeneye_acc_k}. Note that we threshold the accuracy to give a more clear visualization of the most informative area. We also compute a single number accuracy by aggregating the local shared response from all searchlights. This accuracy is compared with accuracy from whole brain factor models with $k=100$ features ($k_1=500,k_2=100$ for 
SR-GICA). The results are shown in Fig.~\ref{fig:acc_bar}.

\begin{figure}[!tb]
\centering
\includegraphics[width=1\linewidth]{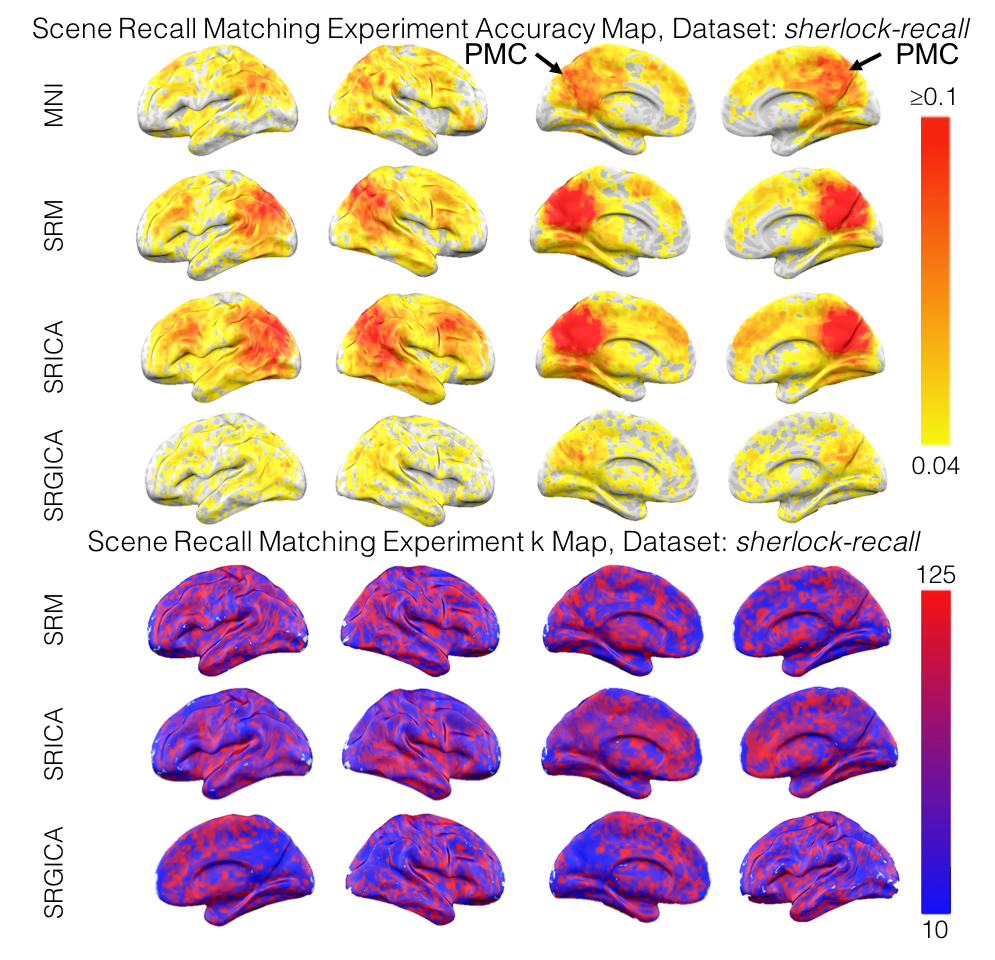}%
\caption{\small Accuracy and k brain maps for scene recall matching experiment using 
{\em sherlock-recall} dataset. PMC: posterior medial cortex.
}\label{fig:recall_acc_k}
\vspace{-3mm}
\end{figure}

\vspace{1mm}
\noindent{\em Scene Recall matching:}
This experiment is designed to test if brain functional patterns are similar when subjects are recalling the same scene. We use the {\em sherlock-movie} 
%and {\em sherlock-recall} datasets where {\em sherlock-movie} is used 
for training and {\em sherlock-recall} for testing. In {\em sherlock-recall}, TRs collected when the subject was recalling the same scene are averaged and projected to the shared response space using $W_i$ learned from the training data. The projected recall data along with the corresponding scene labels are used to train a SVM classifier. The projected recall data from a testing subject is used to test the classifier. Chance accuracy is $0.02$. The accuracy and k brain maps for a subset of models are shown in Fig.~\ref{fig:recall_acc_k}. Accuracies for searchlight factor models and whole brain factor models are computed the same way as time segment matching experiment and are shown in Fig.~\ref{fig:acc_bar}.


\vspace{-2mm}
\section{Discussion and Conclusion} \label{sec:conclusion}
\vspace{-2mm}
We have investigated how well various factor models can locate informative regions 
in a searchlight based analysis of multi-subject fMRI data.
%
%Previous work on multi-subject fMRI data aggregation using factor models shows promising results in aggregating information across the whole brain, but does not show where the information is.
%
This approach highlights local brain regions that are most informative of the cognitive state of interest using both accuracy and $k$ brain maps. 
Early auditory cortex (EAC) and early visual cortex (EVC) are the most informative regions in time segment matching experiment for {\em audiobook} and {\em sherlock-movie} dataset, respectively. This matches the type of stimulus in these datasets. 
Scene recall is a more complex task. In this case a higher level cognitive region, PMC, is more informative.  The results demonstrate that the approach can effectively locate meaningful informative local regions. 
In some neuroscience experiments, it is not clear which regions will be most relevant to the stimulus and/or task. The searchlight factor model approach helps locate regions worthy of further exploration. Moreover, since the searchlight approach preserves spatial locality, we expected the overall accuracy to drop as a consequence of the searchlight constraint. 
In fact, as shown in Fig.~\ref{fig:acc_bar}, the overall accuracy does not drop in most cases, and sometimes even significantly increases. 
%Demonstrates that the quality of the shared information found is not sacrificed. 
The $k$ brain maps also help reveal the effectiveness of the various factor models. 
For example, consider the $k$ brain maps of SR-ICA and SRM. While the accuracy maps of these methods are very close to each other, SR-ICA uses a smaller $k$ to achieve this accuracy. This suggests that each factor in SR-ICA has an improved representation capability. Overall the SR-ICA is showing strong performance across  the three experiments.

\vfill\pagebreak

\bibliographystyle{IEEEbib}
\bibliography{refs}

\end{document}